\pdfoutput=1
\PassOptionsToPackage{dvipsnames}{xcolor}
\documentclass[11pt]{article}

\usepackage[final]{acl}

\usepackage{times}
\usepackage{latexsym}

\usepackage[T1]{fontenc}

\usepackage[utf8]{inputenc}

\usepackage{microtype}

\usepackage{inconsolata}

\usepackage{graphicx}

\usepackage{booktabs}
\usepackage{graphicx}
\usepackage{adjustbox}
\usepackage{amssymb}
\usepackage{makecell}
\usepackage{amsmath}

\usepackage{multirow}
\usepackage{siunitx}
\usepackage{textcomp}

\title{GLiNER-BioMed: A Suite of Efficient Models for Open Biomedical Named Entity Recognition}

\author{
 \textbf{Anthony Yazdani\textsuperscript{1}},
 \textbf{Ihor Stepanov\textsuperscript{2}},
 \textbf{Douglas Teodoro\textsuperscript{1}}
\\
\\
 \textsuperscript{1}Department of Radiology and Medical Informatics,\\Faculty of Medicine, University of Geneva,\\Geneva, Switzerland\\
 \textsuperscript{2}Knowledgator Engineering, Kyiv, Ukraine
\\
 \small{
   \textbf{Correspondence:} \href{mailto:anthony.yazdani@unige.ch}{anthony.yazdani@unige.ch}, \href{mailto:ingvarstep@knowledgator.com}{ingvarstep@knowledgator.com}, \href{mailto:douglas.teodoro@unige.ch}{douglas.teodoro@unige.ch}
 }
}

\begin{document}
\maketitle
\begin{abstract}
Biomedical named entity recognition (NER) presents unique challenges due to specialized vocabularies, the sheer volume of entities, and the continuous emergence of novel entities. Traditional NER models, constrained by fixed taxonomies and human annotations, struggle to generalize beyond predefined entity types. To address these issues, we introduce GLiNER-BioMed, a domain-adapted suite of Generalist and Lightweight Model for NER (GLiNER) models specifically tailored for biomedicine. In contrast to conventional approaches, GLiNER uses natural language labels to infer arbitrary entity types, enabling zero-shot recognition. Our approach first distills the annotation capabilities of large language models (LLMs) into a smaller, more efficient model, enabling the generation of high-coverage synthetic biomedical NER data. We subsequently train two GLiNER architectures, uni- and bi-encoder, at multiple scales to balance computational efficiency and recognition performance. Experiments on several biomedical datasets demonstrate that GLiNER-BioMed outperforms the state-of-the-art in both zero- and few-shot scenarios, achieving 5.96\% improvement in F1-score over the strongest baseline (p-value < 0.001). Ablation studies highlight the effectiveness of our synthetic data generation strategy and emphasize the complementary benefits of synthetic biomedical pre-training combined with fine-tuning on general-domain annotations. All datasets, models, and training pipelines are publicly available at \href{https://github.com/ds4dh/GLiNER-biomed}{https://github.com/ds4dh/GLiNER-biomed}.
\end{abstract}

\section{Introduction}
Named entity recognition (NER) is a key task in biomedical natural language processing, facilitating the automated extraction of entities such as diseases, genes, and chemicals from biomedical texts. As biomedical knowledge evolves, NER models must adapt to emerging terminology, diverse subdomains, and highly specialized vocabularies \cite{park2024biomedical}.

Early biomedical NER systems were primarily rule-based or dictionary-driven, such as MetaMap \cite{aronson2010overview} and cTAKES \cite{savova2010mayo}, relying on structured biomedical ontologies like UMLS \cite{bodenreider2004unified}. These approaches provided high precision for known terms but suffered from low recall and poor generalization to novel or polysemous entities \cite{park2024biomedical}. Statistical methods like conditional random fields improved generalization but required extensive feature engineering and human annotations \cite{lafferty2001conditional, xu2012feature, liu2015feature}. The emergence of transformer-based architectures such as BioBERT \cite{lee2020biobert}, significantly advanced biomedical NER by utilizing contextual embeddings and transfer learning from domain-specific text \cite{lee2020biobert, li2024artificial}. However, the conventional NER approach using pre-trained language models, in which the classification head is fine-tuned using a fixed set of pre-defined entities, limits inference to this entity set \cite{devlin-etal-2019-bert}. As a result, models struggle to generalize beyond predefined labels, limiting their ability to recognize new, domain-specific, or emerging entities \cite{laparra2021review, jolly2024exploring}.

Addressing these issues, \citet{zaratiana-etal-2024-gliner} introduced GLiNER, an efficient encoder-based alternative that leverages natural language label types. GLiNER's key innovation lies in framing NER as a matching problem within a single encoder that jointly represents text and labels, enabling a lightweight and generalizable model for information extraction tasks. GLiNER consistently outperformed generative models like ChatGPT and fine-tuned GPT-style models, operating at a fraction of their parameter size and computational cost.

Despite GLiNER’s promising performance in open NER, directly applying it to biomedical texts remains challenging due to specialized vocabulary, sheer volume of entities, and complex semantic structures unique to biomedical corpora \cite{lee2020biobert, gu2021domain}. Dedicated biomedical adaptation is thus essential. To address this gap, we introduce GLiNER-BioMed, a suite of GLiNER models specifically tailored for biomedicine. Our contributions include:

\begin{itemize}
  \item \textbf{Synthetic data generation:} We developed an efficient, high-throughput model for creating large-scale synthetic biomedical NER datasets.
  \item \textbf{GLiNER-BioMed models:} We pre-trained GLiNER-BioMed models across multiple sizes and architectural variants, ensuring applicability across diverse use cases.
  \item \textbf{Large scale evaluation:} We performed a large-scale evaluation of GLiNER-BioMed across eight biomedical NER benchmarks, demonstrating its effectiveness in identifying a wide range of entity types.
  \item \textbf{Open-source release:} All GLiNER-BioMed and synthetic-annotation models, along with the generated datasets and training pipelines, are publicly accessible at \href{https://github.com/ds4dh/GLiNER-biomed}{https://github.com/ds4dh/GLiNER-biomed}.
\end{itemize}

\section{Related Work}
Recent open NER approaches, also known as universal or generalist NER, have moved beyond fixed taxonomies by reframing entity recognition as reading comprehension or prompt-based tasks. For example, \citet{li-etal-2020-unified} reformulated NER as question answering, while \citet{aly-etal-2021-leveraging} used type descriptions to classify entities from unseen classes. Similarly, \citet{cocchieri-etal-2025-openbioner} proposed OpenBioNER, a BERT-based cross-encoder that injects entity descriptions into token classification to improve generalization to novel biomedical entities. Though significantly smaller, OpenBioNER showed a marginal performance increase over GLiNER-large-v1.0, the first released version of the model.

Parallel efforts have explored generative models for zero- and few-shot NER. The unified information extraction (UIE) framework \cite{lu-etal-2022-unified} unified entity, relation, event, and sentiment extraction into a single text-to-structure generation task. \citet{zhou2023universalner} distilled ChatGPT-generated annotations into UniversalNER (UniNER), achieving strong performance in open NER. In the biomedical domain, \citet{keloth2024advancing} instruction-tuned LLaMA-7B to create BioNER-LLaMA, which matches the performance of specialized NER models. Recently, PP-UIE \cite{paddlenlp2025ppuie} extended UIE using the Qwen2 family of foundation models \cite{yang2024qwen2} for multi-task extraction over longer contexts. Despite these advances, generative approaches continue to face challenges related to computational cost and inference latency \cite{dietrich2024performance}. Moreover, to the best of our knowledge, the BioNER-LLaMA model has not been released to date, limiting reproducibility and comparison against other frameworks.

The GLiNER framework itself, upon which our work builds, has also seen continued development. Subsequent iterations (v2.0, v2.1, v2.5) aimed to improve zero-shot performance. Variants or domain-adapted versions have also been proposed, such as NuNER-Zero, which incorporates annotations generated by large language models (LLMs) for enhanced token and span recognition \cite{bogdanov2024nuner}, and GLiNER-news, optimized specifically for the news domain \cite{tornquist2024curating}.

However, a dedicated biomedical adaptation remained lacking. GLiNER-BioMed addresses this by extending the GLiNER framework for biomedical NER through large-scale synthetic pre-training on biomedical data. Compared to generative models like UniNER or PP-UIE, it offers greater computational efficiency via lightweight span-classification encoders, while providing higher recognition performance.

\section{Method}
To create GLiNER-BioMed, we first constructed a large-scale synthetic pre-training dataset tailored to biomedical NER (Section~\ref{pre_training_dataset}). To further enhance zero-shot generalization, we then leveraged a post-training dataset from the general domain (Section~\ref{post_training_dataset}). Finally, we investigated two architectural variants of GLiNER-BioMed, with different computational complexities, across multiple model scales (Section~\ref{gliner_biomed_models}), each of which underwent pre- and post-training stages.

\subsection{Pre-training dataset}\label{pre_training_dataset}
\begin{figure*}[t]
  \centering
  \includegraphics[width=\textwidth]{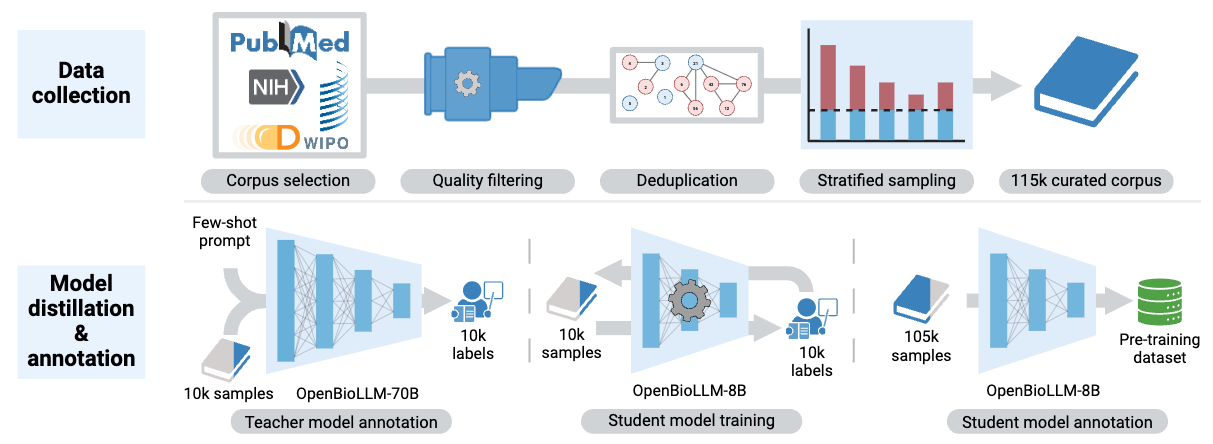}
  \caption{Overview of the synthetic pre-training data generation pipeline for GLiNER-BioMed (Section \ref{pre_training_dataset}). The pipeline begins with data collection, involving corpus selection, quality filtering, deduplication, and stratified sampling to produce a 115k passage corpus. This curated corpus is then annotated using a model distillation strategy where OpenBioLLM-70B (teacher) annotates an initial 10k samples, which are used to train a smaller OpenBioLLM-8B (student) model using low-rank adaptation. The distilled student model then efficiently annotates the remaining 105k passages for the final pre-training dataset.}
  \label{fig:gliner_biomed_dataset}
\end{figure*}

To develop a high-coverage biomedical NER dataset, we first assembled a corpus that encompasses a broad spectrum of biomedical knowledge, including scientific literature (\href{https://pubmed.ncbi.nlm.nih.gov/}{PubMed}), clinical trials (\href{https://clinicaltrials.gov/}{ClinicalTrials.gov}), human prescription labels (\href{https://dailymed.nlm.nih.gov/dailymed/}{DailyMed}), as well as biomedical patents (\href{https://ipcpub.wipo.int/}{World Intellectual Property Organization}). This diverse set of unlabeled biomedical corpora would later be synthetically annotated using a generative LLM. Below, we describe our methodology for corpus selection, quality filtering, deduplication, and synthetic annotation. The complete pipeline is depicted in Figure~\ref{fig:gliner_biomed_dataset}.

\subsubsection{Corpus selection}
PubMed abstracts indexed between January 1 and December 16, 2024, under the MeSH term \textit{pathological conditions, signs, and symptoms} were collected. For clinical trials, we extracted detailed study descriptions as well as arm-level treatment regimens from all trials registered up to November 28, 2024. From DailyMed, we retrieved all available human prescription label sections. Finally, biomedical patent descriptions registered with the WIPO were sourced under the categories A61P (\textit{Specific therapeutic activity of chemical compounds or medicinal preparations}), G16H (\textit{Healthcare informatics}), and A61K (\textit{Preparations for medical, dental or toiletry purposes}) of the International Patent Classification.

\subsubsection{Quality filtering}
We implemented a heuristic-based pipeline to filter low-quality passages. For most corpora, text quality was quantified using a combination of lexical and structural metrics, and passages falling below predefined thresholds were excluded. We applied tailored, more stringent criteria for clinical trial treatment regimen descriptions due to their structural specificity. Details on the metrics and specific filtering rules are provided in Appendix~\ref{app:quality_filtering}.

\subsubsection{Deduplication and content diversity}
To mitigate redundancy, we applied a graph-based deduplication strategy to each biomedical corpus independently. Text passages were first transformed into TF-IDF representations.  We then constructed a similarity graph, where nodes represented individual texts and edges were formed between passages exceeding a cosine similarity of 0.9. Within this graph, connected components emerged as clusters of redundant texts. To retain a maximally informative yet diverse corpus, the passage with the highest average similarity to others was selected as the canonical representative, and all the other instances were excluded. More details are provided in Appendix~\ref{app:deduplication-figure}.

\subsubsection{Stratified sampling}
Next, we performed stratified sampling to construct a balanced dataset for synthetic NER annotation. Prior to sampling, our curated corpus comprised 106,982 PubMed abstracts, 76,145 clinical trial descriptions, 31,778 prescription label sections, 88,867 arm-level treatment regimens, and 114,609 patent descriptions. To optimize computational efficiency while preserving domain diversity, we selected approximately 115,000 passages with equal representation from all sources.

\subsubsection{Model distillation \& annotation}
To construct a large-scale, synthetically annotated NER dataset, we designed a multi-stage process integrating few-shot prompting and model distillation. First, from the balanced subset obtained in the previous step, 10,000 samples were used to create high-quality NER annotations using the OpenBioLLM-70B \cite{OpenBioLLMs} model. In a few-shot setting, incorporating four in-context examples, OpenBioLLM-70B was prompted to annotate these initial 10,000 samples.

To control the scope of annotation, we extracted noun phrases from the input using spaCy \cite{ines_montani_2023_10009823} and provided them to the model as candidate entities. The model was then prompted to classify each noun phrase according to its entity type. This constraint ensured that potentially relevant entities were not omitted. To facilitate parsing, guided decoding was applied to enforce JSON‐formatted outputs \cite{kwon2023efficient}.

Subsequently, we fine-tuned a student model, OpenBioLLM-8B, using low-rank adaptation \cite{hu2022lora} to internalize the annotation behavior of its larger counterpart using these 10,000 annotated samples. This distillation eliminated the need for in-context examples during inference, significantly reducing context length requirements and making annotation more efficient. The student model was then used to annotate the remaining 105,000 samples.

\subsubsection{Synthetic pre-training dataset}
The synthetic pre-training dataset comprises 105,000 samples, 2.3 million entity mentions covering 640,000 unique entities. For exploratory data analysis, 1.5 million of these labels were successfully linked to UMLS concepts via exact string matching, yielding over 2 million CUIs distributed across 120 of the 127 UMLS semantic types and all 15 UMLS semantic groups. This indicates broad coverage of biomedical concepts, encompassing the vast majority of semantic types and all defined semantic groups. Appendix \ref{app:sem-group-breakdown} presents the distribution of synthetic NER labels across UMLS semantic groups.

\subsection{Post-training dataset}\label{post_training_dataset}
To enhance zero-shot performance, we build upon and expand the dataset introduced in \cite{stepanov2024gliner} as the foundation for our post-training data. The base dataset consists of 5,000 instances, combining manually curated examples with synthetic multi-task annotations. Specifically, it includes 1,878 manually curated examples drawn from WNUT2017 \cite{derczynski-etal-2017-results}, OntoNotes5 \cite{hovy-etal-2006-ontonotes}, and MultiNERD \cite{tedeschi-navigli-2022-multinerd}. In addition, it contains 3,122 synthetic examples derived from English Wikipedia articles. These synthetic examples were annotated using Llama-3-8B \cite{grattafiori2024llama}, prompted to perform multiple information extraction (IE) tasks, including NER, open IE, and relation extraction, all formulated as entity recognition tasks to align with the GLiNER framework.

In this work, we substantially augmented this base dataset by generating 14,000 additional synthetic examples, following the same methodology. For this augmentation phase, we used the FineWeb dataset \cite{penedo2024the} as the source and annotated the text using Qwen2.5-72B \cite{yang2024qwen2}. Altogether, the final post-training dataset comprises 19,000 instances, with 337,000 annotated mentions spanning 12,700 unique labels.

To assess the semantic breadth of the NER labels, we mapped entity labels to WordNet, finding that 195,000 labels matched WordNet entries across 37 of all 45 lexnames, and 26 out of 26 noun lexnames. A detailed breakdown of the 15 most frequent lexnames is presented in Appendix~\ref{app:post_wordnet_coverage}.

\subsection{Model architectures and training} \label{gliner_biomed_models}
GLiNER-BioMed employs two distinct model architectures, uni- and bi-encoder, with varied computational complexities. Each architecture was trained across three parameter scales (small, base, and large).

\subsubsection{Uni-encoder architecture}
The standard uni-encoder GLiNER architecture \cite{zaratiana-etal-2024-gliner} uses a single transformer encoder \( f_{\text{enc}} \) to jointly process text and entity types. Given a tokenized text sequence \( T \) and a set of \( k \) target entity types \( E = \{e_i\}_{i=1}^k \), an input sequence \( X = [S_E, T] \) is constructed, where \( S_E \) is a tokenized representation of $E$, constructed by formatting the entity types with special tokens. The encoder processes this combined sequence, generating contextualized hidden states:
$$H = f_{\text{enc}}(X).$$
From the resulting contextualized hidden states \( H \), the text token representations \( H_T \) are extracted. Similarly, a set of \( k \) distinct vector representations for the entity types, \( H_E = \{h_i\}_{i=1}^k \), is also derived from \( H \). These representations are subsequently used to compute span-to-entity-type scores. Specific encoder backbone choices for our implementation are detailed in Appendix~\ref{app:implementation_details_uni_encoder}.

\subsubsection{Bi-encoder architecture}
GLiNER's computational complexity is dominated by the self-attention mechanism over the combined input, which scales as $\mathcal{O}((|S_E| + |T|)^2)$. This becomes prohibitive in settings with a large number of candidate entity types. The bi-encoder model proposed in this work leverages a recently introduced variant of GLiNER \cite{engineering_meet_2024}. This architecture employs two distinct encoders, one for text, $f_{\text{enc}}^{T}$, and one for entity types, $f_{\text{enc}}^{E}$. The text $T$ and entity types $E$ are processed as follows:
$$H_T = f_{\text{enc}}^{T}(T),\quad H_E = \{f_{\text{enc}}^{E}(e_i)\}_{i=1}^k.$$
The label representations \( H_E \) are computed independently of the text and of one another. As a result, they can be pre-computed and cached, offering significant efficiency gains in scenarios involving a large number of entity types. The span-to-entity-type scoring procedure remains the same as in the uni-encoder setup. This separation results in a more favorable complexity profile, with online computation scaling primarily with text length, $\mathcal{O}(|T|^2)$, and a smaller, offline cost for encoding the entity types. Appendix~\ref{app:implementation_details_bi_encoder} specifies the backbones used for each encoder.

\subsubsection{Training procedure}
All model variants underwent pre- and post-training stages. Uni-encoder models were pre-trained exclusively on our fully synthetic biomedical dataset (see Section \ref{pre_training_dataset}). Bi-encoder models, which have a larger parameter count and thus require broader data exposure to converge, were pre-trained on both the NuNER corpus \cite{bogdanov2024nuner} and our synthetic biomedical dataset. In a second stage, all models were fine-tuned on the post-training dataset (see Section~\ref{post_training_dataset}), allowing for broader coverage of general-domain entity types and linguistic contexts. Training hyperparameters are described in Appendix~\ref{app:implementation_details_parameters}.

\section{Evaluation}
We evaluate the performance of GLiNER-BioMed models in both zero-shot and few-shot settings. To ensure a comprehensive assessment, we compute three performance metrics: (1) micro F1-score, which provides an overall measure of precision-recall balance; (2) macro mean F1-score, which averages F1-scores per entity type, treating all entity categories equally regardless of their frequency; and (3) macro median F1-score, which reports the median F1-score across all entity types, mitigating the influence of outliers.

For statistical comparisons, we assess the differences between two models using a Wilcoxon signed-rank test on their per-passage F1-scores, specifically testing whether Model A’s F1-scores exceed Model B’s. A low p-value provides significant evidence that Model A outperforms Model B; a high p-value implies there is insufficient evidence for that claim.

\subsection{Benchmark datasets}
For the evaluation, we consider eight publicly available, human-annotated NER benchmarks covering a broad range of biomedical subdomains, including clinical narratives, scientific abstracts, regulatory documents, drug labels, and patient-generated content. These include TAC \cite{roberts2017overview}, CADEC \cite{karimi2015cadec}, N2C2 2018 \cite{henry20202018}, BC5CDR \cite{li2016biocreative}, BioRED \cite{luo2022biored}, CHIA \cite{kury2020chia}, Biomed NER \cite{knowledgatorbiomed_ner_2025}, and NCBI Disease \cite{dougan2014ncbi}. In total, these datasets comprise 10,918 text passages and 85,959 entity mentions spanning 58 unique biomedical entity types. Dataset descriptions and preprocessing details are provided in Appendix~\ref{app:benchmark_datasets}.

\subsection{Results on zero-shot performance}
We evaluate GLiNER-BioMed models in zero-shot settings against a range of GLiNER-based and generative baselines, as shown in Table~\ref{tab:model_comparison}. The GLiNER baselines include the general-purpose versions v1.0 \cite{zaratiana-etal-2024-gliner}, v2.0, v2.1, and v2.5, as well as NuNER-Zero, NuNER-Zero-span \cite{bogdanov2024nuner}, and GLiNER-news-v2.1 \cite{tornquist2024curating}. The generative baselines include biomedical chat models, namely, OpenBioLLM in 8B and 70B; general-domain chat models, Qwen3 in 8B and 32B \cite{qwen3}; and five dedicated information extraction models, including UniNER-7B \cite{zhou2023universalner} and PP-UIE in 0.5B, 1.5B, 7B, and 14B configurations \cite{paddlenlp2025ppuie}.

\begin{table}[htbp]
    \centering
    \setlength{\tabcolsep}{4pt}
    \renewcommand{\theadfont}{\small\bfseries}
    \begin{tabular}{l S[table-format=2.2]
                     S[table-format=2.2]
                     S[table-format=2.2]}
        \toprule
        \thead{Model} & {\thead{Micro\\F1}} & {\thead{Macro\\Mean\\F1}} & {\thead{Macro\\Median\\F1}} \\ 
        \midrule
        \multicolumn{4}{l}{\thead{\textbf{Generative LLMs}}} \\ \midrule
        OpenBioLLM-70B\textsuperscript{$*$} & 36.30 & 20.86 & 14.40 \\
        Qwen3-32B\textsuperscript{$\dagger$}& \underline{37.67} & \underline{26.55} & \underline{22.27} \\
        PP-UIE-14B                          & 25.74 & 18.05 & 13.73 \\
        OpenBioLLM-8B                       & 15.95 &  9.29 &  3.95 \\
        Qwen3-8B\textsuperscript{$\dagger$} & 34.04 & 21.47 & 10.84 \\
        UniNER-7B                           & \textbf{48.55} & \textbf{30.35} & \textbf{28.10} \\
        PP-UIE-7B                           & 24.73 & 17.30 & 13.25 \\
        PP-UIE-1.5B                         & 23.87 & 16.30 & 11.59 \\
        PP-UIE-0.5B                         & 19.78 & 13.27 &  9.27 \\
        \midrule
        \multicolumn{4}{l}{\thead{\textbf{GLiNER models}}} \\ \midrule
        \multicolumn{4}{l}{\textbf{Large}} \\ \midrule 
        NuNER-Zero                          & 40.87 & 21.79 & 13.94 \\
        NuNER-Zero-span                     & 40.26 & 22.51 & 14.27 \\
        GLiNER-v1.0                         & 47.77 & 29.60 & 21.13 \\
        GLiNER-v2.0                         & 37.38 & 21.42 & 15.44 \\
        GLiNER-v2.1                         & 48.04 & 29.75 & 28.20 \\
        GLiNER-news-v2.1                    & 48.99 & 31.79 & 33.77 \\
        GLiNER-v2.5                         & 53.81 & 35.22 & \underline{35.65} \\
        \midrule
        GLiNER-BioMed                       & \textcolor{ForestGreen}{\textbf{59.77}} & \textcolor{ForestGreen}{\textbf{40.67}} & \textcolor{ForestGreen}{\textbf{42.65}} \\
        GLiNER-BioMed-bi                    & \underline{54.90} & \underline{35.78} & 31.66 \\
        \midrule
        \multicolumn{4}{l}{\textbf{Base}} \\ \midrule 
        GLiNER-v1.0                         & 41.61 & 24.98 & 10.27 \\
        GLiNER-v2.0                         & 34.33 & 24.48 & 22.01 \\
        GLiNER-v2.1                         & 40.25 & 25.26 & 14.41 \\
        GLiNER-news-v2.1                    & 41.59 & 27.16 & 17.74 \\
        GLiNER-v2.5                         & 46.49 & 30.93 & 25.26 \\
        \midrule
        GLiNER-BioMed                       & \underline{54.37} & \textbf{36.20} & \textbf{41.61} \\
        GLiNER-BioMed-bi                    & \textbf{58.31} & \underline{35.22} & \underline{32.39} \\
        \midrule
        \multicolumn{4}{l}{\textbf{Small}} \\ \midrule 
        GLiNER-v1.0                         & 40.99 & 22.81 &  7.86 \\
        GLiNER-v2.0                         & 33.55 & 21.12 & 15.76 \\
        GLiNER-v2.1                         & 38.45 & 23.25 & 10.92 \\
        GLiNER-news-v2.1                    & 39.15 & 24.96 & 14.48 \\
        GLiNER-v2.5                         & 38.21 & 28.53 & 18.01 \\
        \midrule
        GLiNER-BioMed                       & \underline{52.53} & \textbf{34.49} & \textbf{38.17} \\
        GLiNER-BioMed-bi                    & \textbf{56.93} & \underline{33.88} & \underline{33.61} \\
        \bottomrule
    \end{tabular}
    \caption{Zero-shot NER performance of GLiNER-BioMed models against 17 GLiNER baselines across three model sizes (large, base, small) and 9 generative LLMs, aggregated over eight datasets. Bold: best in size category; underlined: second-best; green: overall best. \textsuperscript{$*$}Q4KM-quantized. \textsuperscript{$\dagger$}Thinking mode disabled.}
    \label{tab:model_comparison}
\end{table}

As shown in Table~\ref{tab:model_comparison}, GLiNER-BioMed outperforms all baselines across model sizes. In the large model category, it achieves a micro F1-score of 59.77\%, improving by 5.96 points over the next-best model, GLiNER-v2.5-large ($p\!<\!0.001$). Notably, despite having seven times fewer parameters, GLiNER-BioMed-small achieves performance comparable to GLiNER-v2.5-large, with no statistical evidence of superior performance ($p\!>\!0.05$).

Against generative LLMs, GLiNER-BioMed-large also delivers stronger zero-shot performance. It surpasses UniNER-7B, the strongest generative baseline, by 11.22 points ($p\!<\!0.001$), and outperforms the top-performing PP-UIE model, PP-UIE-14B, by a wide margin ($p\!<\!0.001$). OpenBioLLM-70B and OpenBioLLM-8B reach F1-scores of 36.30\% and 15.95\%, respectively. These results show that GLiNER-BioMed not only surpasses state-of-the-art baselines but also the foundation models used to annotate its training data ($p\!<\!0.001$ in both cases).

GLiNER-BioMed-bi models show distinct behavior. At the small scale, it achieves a 56.93\% F1-score, outperforming the uni-encoder by 4.40 points ($p\!<\!0.001$). Similarly, at the base scale, it scores 58.31\%, an improvement of 3.94 points over its uni-encoder counterpart ($p\!<\!0.001$). However, at the large scale, its performance is 4.87 points lower than that of the uni-encoder ($p\!<\!0.001$). We hypothesize that the bi-encoder framework improves performance in small and base models by doubling encoder capacity, mitigating their inherent parameter limitations. However, at larger scales, this advantage fades, as uni-encoder architectures might already have sufficient capacity.

\subsection{Results on few-shot performance}\label{sec:few_shot_eval}
Few-shot learning is particularly valuable in biomedical contexts, where annotated data is often limited due to the high cost and domain expertise required for manual labeling. To assess model performance under such constrained supervision, we fine-tune GLiNER-BioMed-large and GLiNER-BioMed-bi-large on 10, 20, and 50-shot subsets drawn from the training splits of each dataset, comparing them to the strongest baseline, GLiNER-v2.5-large. All models are evaluated on full test sets for consistency with zero-shot results.

\begin{table}[htbp]
    \centering
    \begin{tabular}{l S[table-format=2.2] 
                     S[table-format=2.2]
                     S[table-format=2.2]}
        \toprule
        \textbf{N-shot} & {\textbf{v2.5}} & {\textbf{BioMed}} & {\textbf{BioMed-bi}} \\
        \midrule
        \hphantom{0}0-shot         & 53.81          & \textbf{59.77} & \underline{54.90}   \\
        \midrule
        10-shot        & 65.93          & \underline{66.07} & \textbf{70.39}   \\
        20-shot        & 69.15          & \underline{71.98} & \textbf{73.07}   \\
        50-shot        & 73.52          & \underline{73.70} & \textbf{76.02}   \\
        \midrule
        Full dataset   & 84.64          & \textbf{84.95}    & \underline{84.91}   \\
        \bottomrule
    \end{tabular}
    \caption{Few-shot NER performance of GLiNER-BioMed large models vs. GLiNER-v2.5-large, aggregated over eight datasets using micro F1-scores. N-shot denotes N training and validation samples. Bold: best per setting; underlined: second-best.}
    \label{tab:few_shot_micro_f1}
\end{table}

As shown in Table~\ref{tab:few_shot_micro_f1}, GLiNER-BioMed-bi-large consistently outperforms the v2.5 and uni-encoder versions in the few-shot setting, reaching a 70.39\% micro F1-score with only 10 examples, representing a 4.46 point improvement over GLiNER-v2.5-large ($p\!<\!0.001$). This advantage over GLiNER-v2.5-large holds at 20 and 50-shot, with $p\!<\!0.05$ and $p\!<\!0.001$, respectively. The uni-encoder GLiNER-BioMed-large also performs well in these N-shot scenarios, slightly outperforming GLiNER-v2.5-large, although not statistically significantly ($p\!>\!0.05$ for all comparisons).

Under full supervision, all models converge. In this setting, both GLiNER-BioMed variants perform comparably to GLiNER-v2.5-large, with $p\!>\!0.05$ for all comparisons. This result suggests that the uni-encoder GLiNER-BioMed is best suited for zero-shot scenarios, while the bi-encoder variant excels in annotation-constrained settings, improving performance upon zero-shot as much as 15.49 points ($p\!<\!0.001$), with only 10 annotated samples.

\subsection{Uni- vs. bi-encoder inference efficiency}

\begin{figure*}[t]
  \centering
  \includegraphics[width=\textwidth]{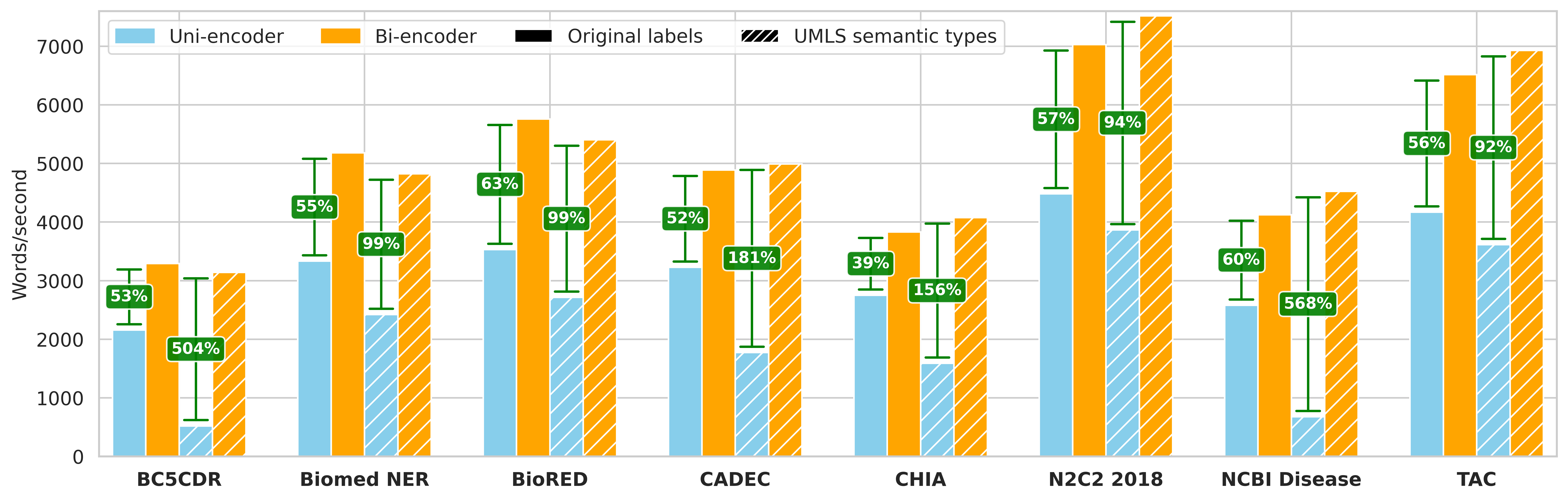}
  \caption{Peak inference throughput (words/second) for GLiNER-BioMed large models, illustrating architectural efficiency under varied entity type loads. Peak words/second is the maximum achieved over batch sizes going from 1 to 64 on an NVIDIA RTX3090 GPU before out-of-memory using FP32 precision. Performance of uni-encoder (blue) and bi-encoder (orange) is compared using (1) dataset-specific entity labels (solid bars), and (2) a fixed set of 127 UMLS semantic types (hatched bars). Annotated percentages indicate the throughput difference between bi- and uni-encoders.}
  \label{fig:grouped_peak_throughput_large}
\end{figure*}

The practical utility of NER models critically depends on inference efficiency. To evaluate this, we benchmarked the large GLiNER-BioMed and GLiNER-BioMed-bi variants on an NVIDIA RTX3090 GPU, measuring peak throughput in words per second using FP32 precision. Evaluations were conducted under two conditions: (1) using the original set of entity types for each dataset, and (2) using a set comprising all 127 UMLS semantic types. As shown in Figure~\ref{fig:grouped_peak_throughput_large}, the bi-encoder outperformed the uni-encoder in throughput in both settings. With dataset-specific labels, it was between 39\% to 63\% faster. This performance gap widened substantially under the full UMLS label set, where the uni-encoder processes text and labels jointly through self-attention, causing inference time to grow quadratically with the combined sequence length. In contrast, the bi-encoder decouples text and label encoding, allowing label embeddings to be pre-computed and cached, drastically reducing inference overhead. Under the 127-label setting, this resulted in speedups ranging from 92\% to 568\%, depending on the dataset.

\section{Ablation studies}
We perform ablation studies to quantify the individual and combined contributions of synthetic biomedical pre-training and general-domain post-training. Results are summarized in Table~\ref{tab:ablation_micro_f1}.

\begin{table*}[t]
    \centering
    \begin{tabular}{ccc S[table-format=2.2]
                         S[table-format=2.2]
                         S[table-format=2.2]}
        \toprule
        \multicolumn{3}{c}{\textbf{Training phases}} & {\multirow{2}{*}[-10pt]{{\textbf{Precision}}}} & {\multirow{2}{*}[-10pt]{{\textbf{Recall}}}} & {\multirow{2}{*}[-10pt]{{\textbf{F1-score}}}} \\ 
        \cmidrule(lr){1-3}
        \makecell{General-domain \\ pre-training} & \makecell{Pre-training \\ (Section~\ref{pre_training_dataset})} & \makecell{Post-training \\ (Section~\ref{post_training_dataset})} & & & \\ 
        \midrule
        \checkmark        & \texttimes          & \texttimes          & 56.19              & 51.62             & 53.81            \\
        \checkmark        & \texttimes          & \checkmark          & 55.56              & 54.06             & 54.80            \\
        \texttimes        & \texttimes          & \checkmark          & 51.53              & 53.25             & 52.38            \\
        \texttimes        & \checkmark          & \texttimes          & \textbf{70.08}     & 30.09             & 42.10            \\
        \texttimes        & \checkmark          & \checkmark          & 56.67              & \textbf{63.22}    & \textbf{59.77}   \\
        \bottomrule
    \end{tabular}
    \caption{Ablation study evaluating the impact of different training phases on biomedical NER performance. The evaluated model configurations are: (1) GLiNER-v2.5-large trained solely on general-domain data (baseline); (2) GLiNER-v2.5-large trained on general-domain data and further trained on the post-training data (Section~\ref{post_training_dataset}); (3) randomly initialized GLiNER-large trained only on the post-training data; (4) randomly initialized GLiNER-large trained exclusively on synthetic biomedical data (Section~\ref{pre_training_dataset}); and (5) randomly initialized GLiNER-large trained on our synthetic biomedical data and subsequently trained on the post-training data (GLiNER-BioMed-large). All reported metrics are micro-averaged.}
    \label{tab:ablation_micro_f1}
\end{table*}

The baseline model, GLiNER-v2.5-large, trained solely on general-domain data, achieves 53.81\% micro F1-score. Adding post-training yields a modest gain to 54.80\%, suggesting some benefit from high-quality annotations but limited domain adaptation. Training a randomly initialized model on post-training data alone results in a similar F1-score of 52.38\%, confirming the quality of the data while highlighting the need for domain-specific exposure.

Training a model exclusively on synthetic biomedical data yields 42.10\% F1-score, with high precision (70.08\%) but low recall (30.09\%). This result indicates that synthetic biomedical pre-training effectively imparts domain-specific knowledge but lacks sufficient recall to achieve the highest F1-score. The best performance comes from combining biomedical synthetic pre-training with general-domain post-training, resulting in 59.77\% F1-score, with balanced precision (56.67\%) and recall (63.22\%).

These results demonstrate the effectiveness of using cost-efficient generative models to distill biomedical knowledge into GLiNER, and show that fine-tuning on high-quality, diverse data is critical for maximizing recall and overall recognition performance.

\section{Conclusion}
This work introduces GLiNER-BioMed, a specialized suite of open biomedical NER models. Unlike conventional approaches that rely on fixed taxonomies, GLiNER-BioMed incorporates natural language labels, enabling more flexible and adaptive entity recognition. Our approach begins by distilling annotations from a large generative biomedical model into a smaller generative model, which is then used to generate a high-coverage synthetic biomedical NER dataset. We then pre-train GLiNER models on this synthetic dataset, followed by post-training on high-quality general-domain data, enhancing annotation accuracy while preserving strong domain adaptation.

Through extensive zero-shot and few-shot evaluations across eight biomedical datasets, GLiNER-BioMed consistently outperforms state-of-the-art information extraction models, achieving a 5.96-point F1-score improvement over the strongest baseline. Additionally, GLiNER-BioMed-bi, the bi-encoder variant, proves particularly effective in low-data settings, achieving a 70.39\% micro F1-score with as few as 10 annotated samples and further improving to 76.02\% with 50 annotated instances. Beyond accuracy, GLiNER-BioMed-bi offers substantial efficiency gains, achieving 39\% to 568\% higher throughput compared to the uni-encoder across a range of evaluation scenarios. These results demonstrate its advantage in few-shot learning and resource-constrained scenarios, highlighting its potential for real-world biomedical applications.

Future research could explore more capable and scalable generative large language models for synthetic data annotations, multilingual adaptations, and continual learning strategies to ensure robust adaptation to the ever-evolving biomedical landscape.

\section*{Limitations}
Although GLiNER-BioMed achieves substantial gains, several limitations remain. First, our synthetic pre-training data, generated via distilled generative models, may introduce biases or fail to fully capture the complexity inherent in human-annotated data, potentially limiting generalizability. Additionally, despite their diversity, the evaluation datasets may not represent all biomedical subdomains or linguistic variations, especially less common areas, such as veterinary medicine or dentistry. Furthermore, despite optimization efforts, the computational resources required by our complete pipeline may pose barriers to reproducibility and adoption for researchers operating in resource-constrained environments. Finally, while quantitatively extensive, our evaluation currently lacks detailed qualitative analysis, limiting deeper insights into model errors, interpretability, and possible directions for further improvement.

\bibliography{custom}

\appendix

\setcounter{figure}{0}
\setcounter{table}{0}

\renewcommand{\thefigure}{S\arabic{figure}}
\renewcommand{\thetable}{S\arabic{table}}

\section{Text quality filtering details} \label{app:quality_filtering}

We applied a set of heuristic quality filters to human prescription labels, clinical trial detailed descriptions, PubMed abstracts, and biomedical patents. Texts were excluded if they failed any of our quality thresholds. Specifically, we required passages to contain no more than 30\% words with non-alphabetic characters, at least 6 sentences, and an average sentence length of 10 words or more. The proportion of uppercase letters among all alphabetic characters was limited to a maximum of 20\%, and no single word could exceed 20\% of the total word count to avoid excessive repetition. Passages were also removed if they contained too many line breaks, defined as more than one newline cluster for every two sentences. Additionally, we ensured at least 10\% lexical diversity, meaning at least 10\% of the words needed to be unique, and required a minimum of 5\% common English stopwords.

Due to their distinctive formatting, we applied more stringent criteria for filtering clinical trial treatment regimens. Passages were retained only if they began with a capital letter, included at least one digit, ended with a period, and contained at least two well-formed sentences.

\section{Graph-based semantic deduplication} \label{app:deduplication-figure}

\begin{figure}[h]
  \centering
  \includegraphics[width=\columnwidth]{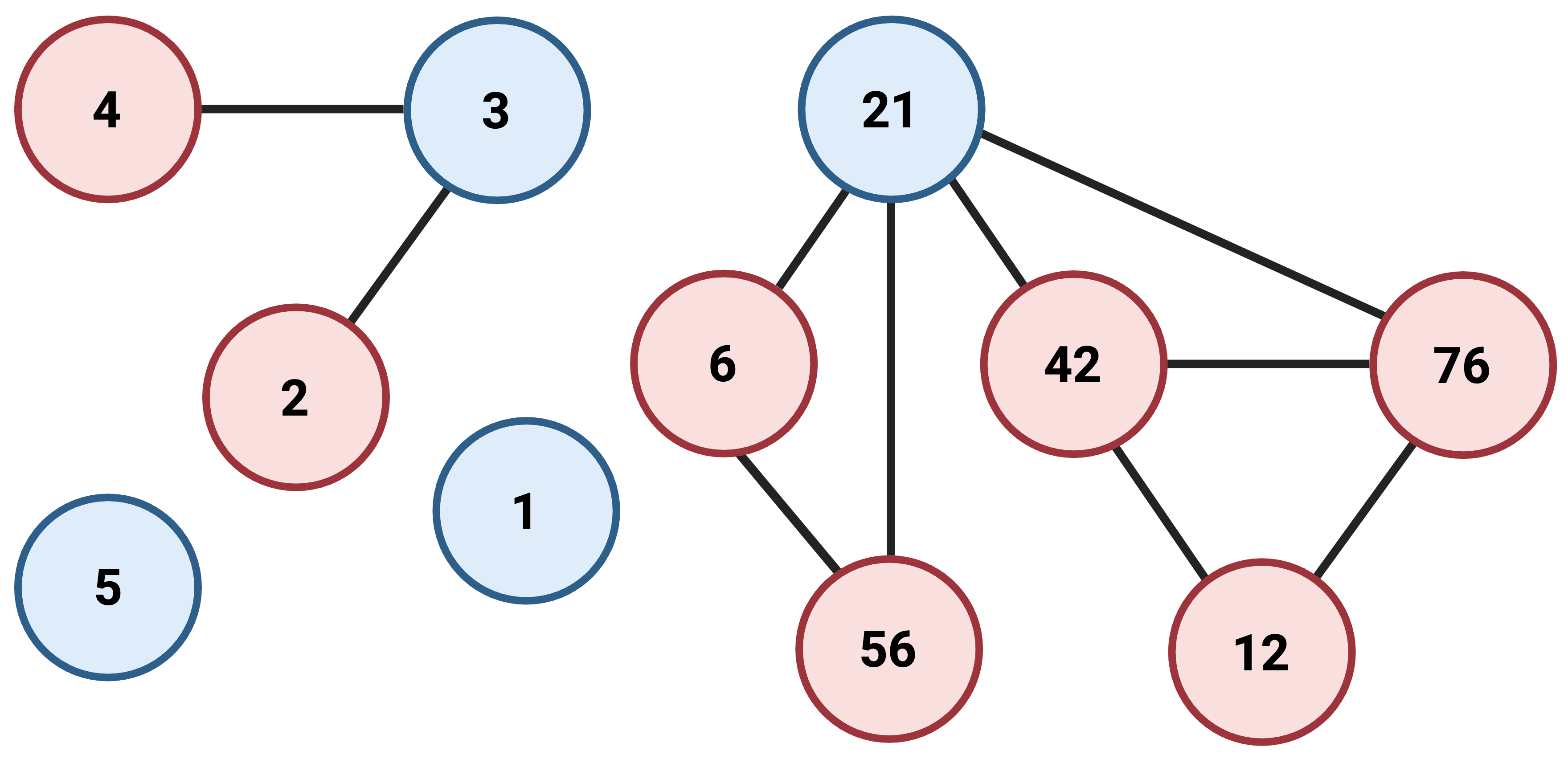}
  \caption{TF-IDF similarity graph for biomedical passages. Blue nodes represent retained representatives; red nodes indicate excluded duplicates.}
  \label{fig:semantic_deduplication_gliner_biomed}
\end{figure}

To illustrate our deduplication strategy, Figure~\ref{fig:semantic_deduplication_gliner_biomed} shows an example of a TF-IDF-based similarity graph constructed from biomedical passages. In this graph, each node represents a text passage, and edges connect pairs of passages whose cosine similarity exceeds 0.9. Connected components reflect clusters of redundant texts. For each cluster, we retain only the passage with the highest average similarity to the others, discarding the rest to ensure diversity without sacrificing coverage.

\section{Entity type distribution in pre-training dataset} \label{app:sem-group-breakdown}

\begin{figure}[h]
  \centering
  \includegraphics[width=\columnwidth]{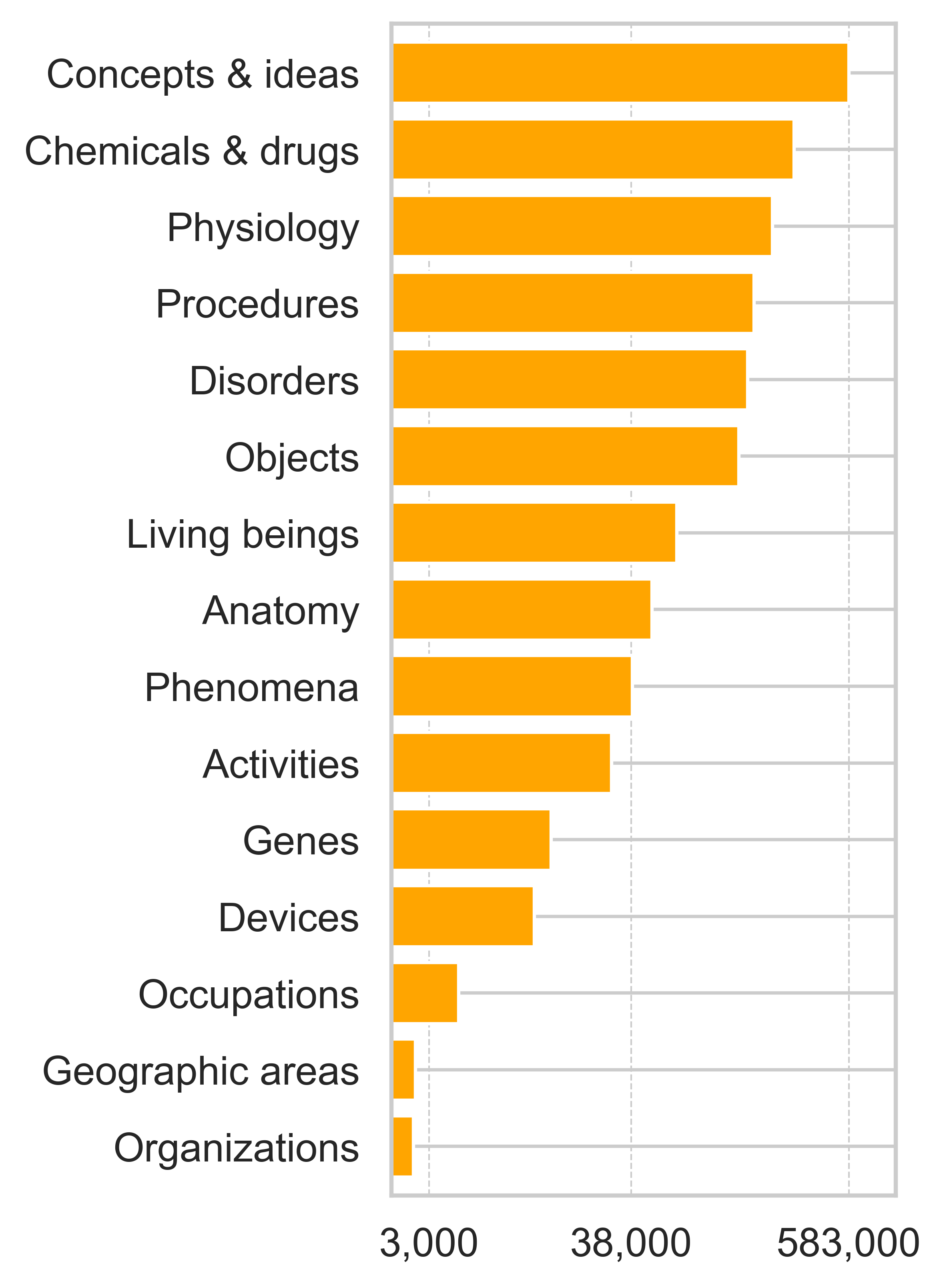}
  \caption{UMLS semantic groups by mention count (log scale) in the synthetic pre-training dataset. Counts reflect total NER label–to–semantic group mappings; labels associated with multiple semantic groups contribute to multiple bars.}
  \label{fig:semantic_group_distribution_neperian_log}
\end{figure}

Figure~\ref{fig:semantic_group_distribution_neperian_log} presents the distribution of all 15 UMLS semantic groups found in the synthetic pre-training dataset. This breakdown offers insight into the biomedical concept coverage of the pre-training dataset, highlighting the most commonly represented entity types.

\section{WordNet coverage in post-training dataset} \label{app:post_wordnet_coverage}

\begin{figure}[h]
  \centering
  \includegraphics[width=\columnwidth]{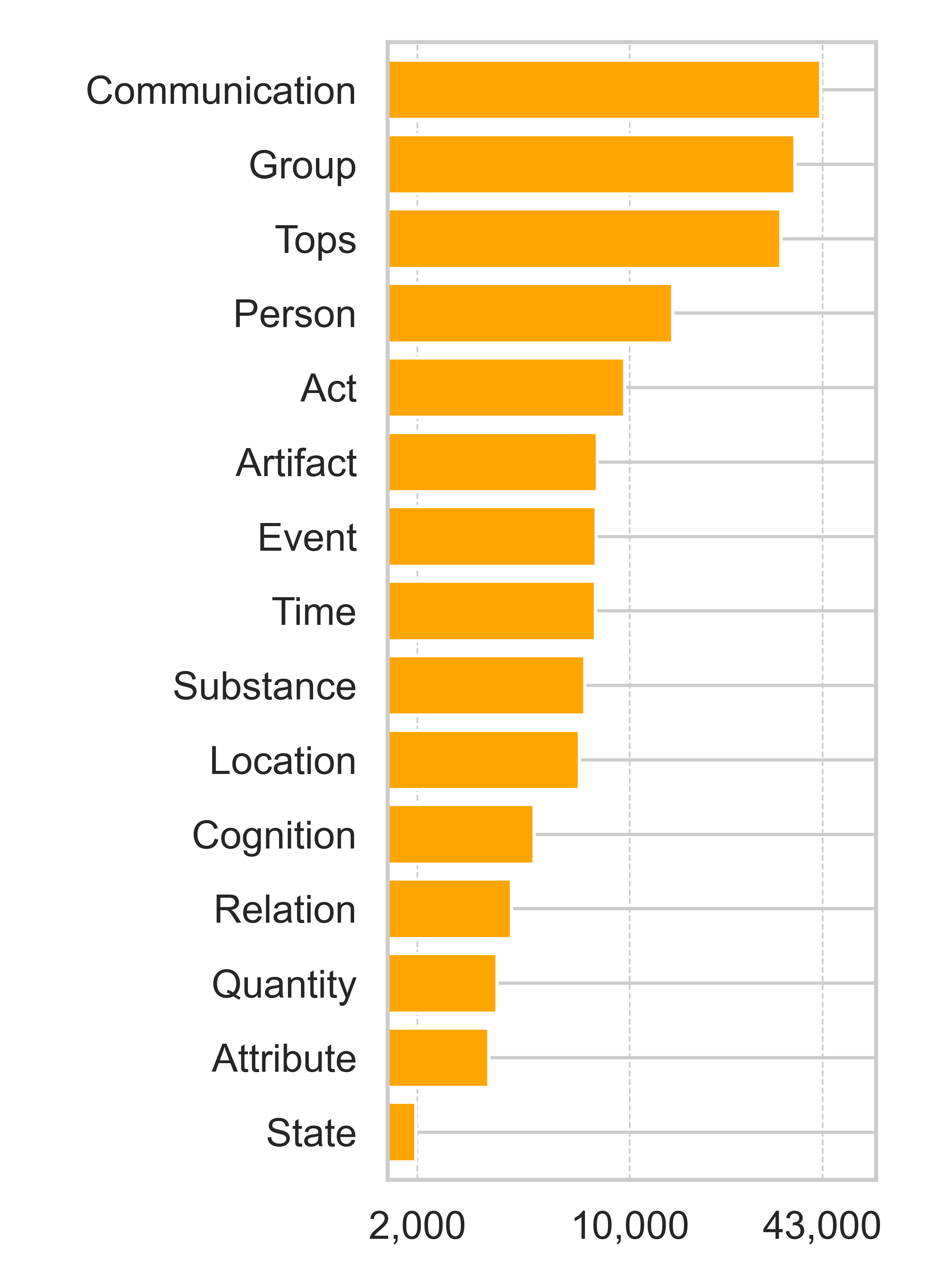}
  \caption{Top 15 WordNet lexnames by mention count (log scale) in the post-training dataset. Each mention is assigned a single lexname based on its most frequent WordNet sense.}
  \label{fig:wordnet_lexname_distribution_styled_log_final}
\end{figure}

Figure~\ref{fig:wordnet_lexname_distribution_styled_log_final} shows the distribution of the 15 most frequent WordNet lexnames in the post-training dataset. Lexnames are coarse semantic categories used by WordNet to group concepts. We mapped each entity label to the first WordNet synset, corresponding to its most frequent sense, and assigned the associated lexname. Unlike the pre-training dataset, where each label could map to multiple UMLS semantic groups, WordNet provides a frequency-based ordering of senses, allowing us to select a single most likely interpretation.

\section{Implementation details} \label{app:implementation_details}

\subsection{Uni-encoder GLiNER-BioMed} \label{app:implementation_details_uni_encoder}
For the uni-encoder variants of GLiNER-BioMed, we used \textit{DeBERTa-v3} backbones \cite{he2021debertav3}, specifically, the small, base, and large models, as the encoder component \( f_{\text{enc}} \).

\subsection{Bi-encoder GLiNER-BioMed} \label{app:implementation_details_bi_encoder}
For the bi-encoder architecture, $f_{\text{enc}}^{T}$ was \textit{DeBERTa-v3} (small, base, large). The label encoder $f_{\text{enc}}^{E}$ utilized \textit{all-MiniLM-L6-v2} \cite{reimers-2019-sentence-bert} for small, \textit{bge-small-en-v1.5} \cite{bge_embedding} for base, and \textit{bge-base-en-v1.5} \cite{bge_embedding} for large.

\subsection{Training hyperparameters} \label{app:implementation_details_parameters}
Both the pre- and post-training datasets were split into 90\% training and 10\% validation sets. During synthetic biomedical pre-training, models were trained for 20,000 steps with a batch size of 8, using the AdamW optimizer \cite{loshchilov2017decoupled} with a weight decay of 0.01. The learning rate was set to $1 \times 10^{-5}$ for the encoder backbone and $5 \times 10^{-5}$ for all other model parameters. For post-training, models were trained for 10,000 steps with a batch size of 4, again using AdamW with a weight decay of 0.01. The learning rate during this phase was reduced to $5 \times 10^{-6}$ for the encoder backbone and $1 \times 10^{-5}$ for the remaining parameters.

\section{Benchmark datasets details and preprocessing} \label{app:benchmark_datasets}

\subsection{NER benchmark datasets}
We evaluate GLiNER-BioMed on eight biomedical NER datasets spanning a broad range of entity types. TAC 2017 \cite{roberts2017overview} consists of structured drug labels annotated with adverse drug event (ADE) mentions and contextual cues such as severity and negation. CADEC \cite{karimi2015cadec} contains user forum posts related to medications and health issues, reflecting informal language and covering ADEs, symptoms, and drug mentions. N2C2 2018 \cite{henry20202018} focuses on medication-related concepts in discharge summaries, such as drug, dosage, frequency, route, duration, ADEs, and indications. BC5CDR \cite{li2016biocreative} is a widely used benchmark for recognizing chemical and disease mentions in PubMed abstracts. BioRED \cite{luo2022biored} broadens the biomedical NER scope to include genes, sequence variants, organisms, and cell lines in scientific abstracts. CHIA \cite{kury2020chia} covers clinical trial eligibility criteria, with annotations spanning conditions, procedures, devices, and other medical concepts. BioMed NER \cite{knowledgatorbiomed_ner_2025} is a large-scale dataset covering a broad spectrum of biomedical, clinical, and regulatory entity types, including drugs, anatomical structures, phenotypes, and legal concepts. Finally, NCBI Disease \cite{dougan2014ncbi} is a curated dataset for disease name recognition in biomedical abstracts. An overview of total and test mention counts for each dataset is provided in Table~\ref{tab:benchmark_datasets}.

\begin{table}[h]
\centering
\begin{tabular}{lcc}
\toprule
\textbf{Dataset} & \multicolumn{2}{c}{\textbf{Mention counts}} \\
\cmidrule(lr){2-3}
 & Total & Test \\
\midrule
BC5CDR         & 28,785  & 9,809  \\
BioMed NER     & 190,330 & 19,314 \\
BioRED         & 20,419  & 3,535  \\
CADEC          & 8,045   & 1,302  \\
CHIA           & 47,081  & 4,745  \\
N2C2 2018      & 83,869  & 32,918 \\
NCBI Disease   & 6,892   & 960    \\
TAC            & 28,122  & 13,478 \\
\bottomrule
\end{tabular}
\caption{Overview of benchmark datasets used for evaluation. Mention counts refer to annotated entity spans. Test mentions refer to entities in the held-out test set.}
\label{tab:benchmark_datasets}
\end{table}

\subsection{Preprocessing details}
To ensure consistency across all benchmark datasets, we applied a unified preprocessing pipeline. All texts were tokenized using spaCy, and entity annotations, originally provided in different formats such as BRAT, BIO tags, or XML, were mapped to token-level start and end indices. Where available, we retained official train/val/test splits; otherwise, we created stratified splits using fixed random seeds for reproducibility. Discontinuous entities were excluded to ensure compatibility with the GLiNER framework, and entity type names were standardized or renamed to ensure self-contained and interpretable label sets. To handle long documents, we chunked examples into segments capped at 512 subword tokens using the DeBERTa tokenizer, ensuring that no entity span was split across chunks. Each chunk was treated as an independent instance, and missing entity types were recorded as negatives to support learning from absence during few-shot and full-data training.

\end{document}